\documentclass[runningheads]{llncs}
\usepackage{graphicx}

\usepackage{tikz}
\usepackage{comment}
\usepackage{amsmath,amssymb} 
\usepackage{color}
\usepackage{tabularx}
\usepackage{multirow}
\usepackage{booktabs}
\usepackage{subfigure}
\usepackage{enumerate}
\usepackage[hidelinks]{hyperref}

\usepackage[accsupp]{axessibility}  

\newcommand{\etc}{\textit{etc.}}

\newcommand{\gt}{\textit{gt}}
\newcommand{\posneg}{\textit{pos/neg}}

\newcommand{\equspace}{4pt}

\newcolumntype{x}{>\small c} 
\newcolumntype{L}[1]{>{\raggedright\let\newline\\\arraybackslash\hspace{0pt}}m{#1}} 
\newcolumntype{C}[1]{>{\centering\let\newline\\\arraybackslash\hspace{0pt}}m{#1}} 
\newcolumntype{R}[1]{>{\raggedleft\let\newline\\\arraybackslash\hspace{0pt}}m{#1}}

\begin{document}
\pagestyle{headings}
\mainmatter

\title{RFLA: Gaussian Receptive Field based Label Assignment for Tiny Object Detection}

\titlerunning{RFLA}
%
\author{Chang Xu\inst{1} \and
Jinwang Wang\inst{2} \and
Wen Yang\inst{1}\thanks{Corresponding Author} \and
Huai Yu\inst{1} \and
Lei Yu\inst{1} \and
Gui-Song Xia\inst{3}}
\authorrunning{C. Xu et al.}
%
\institute{School of Electronic Information, Wuhan University  \and
Huawei Technologies Co., Ltd.  \and
School of Computer Science, Wuhan University \\
\email{\{xuchangeis,yangwen,yuhuai,ly.wd,guisong.xia\}@whu.edu.cn} \email{wangjinwang3@huawei.com}}

\maketitle

\begin{abstract}

Detecting tiny objects is one of the main obstacles hindering the development of object detection. The performance of generic object detectors tends to drastically deteriorate on tiny object detection tasks. In this paper, we point out that either box prior in the anchor-based detector or point prior in the anchor-free detector is sub-optimal for tiny objects. Our key observation is that the current anchor-based or anchor-free label assignment paradigms will incur many outlier tiny-sized ground truth samples, leading to detectors imposing less focus on the tiny objects. To this end, we propose a Gaussian Receptive Field based Label Assignment (RFLA) strategy for tiny object detection. Specifically, RFLA first utilizes the prior information that the feature receptive field follows Gaussian distribution. Then, instead of assigning samples with IoU or center sampling strategy, a new Receptive Field Distance (RFD) is proposed to directly measure the similarity between the Gaussian receptive field and ground truth. Considering that the IoU-threshold based and center sampling strategy are skewed to large objects, we further design a Hierarchical Label Assignment (HLA) module based on RFD to achieve balanced learning for tiny objects. Extensive experiments on four datasets demonstrate the effectiveness of the proposed methods. Especially, our approach outperforms the state-of-the-art competitors with 4.0 AP points on the AI-TOD dataset. Codes are available at \url{https://github.com/Chasel-Tsui/mmdet-rfla}.

\end{abstract}

\section{Introduction}

Tiny object, featured by its extremely limited amount of pixels (less than 16 $\times$ 16 pixels defined in AI-TOD~\cite{AI-TOD_2020_ICPR}), is always \textit{a hard nut to crack} in the computer vision community. Tiny Object Detection (TOD) is one of the most challenging tasks, and generic object detectors usually fail to provide satisfactory results on TOD tasks~\cite{AI-TOD_2020_ICPR,TinyPerson_2020_WACV}, resulting from tiny object's lack of discriminative features. Considering the particularity of tiny objects, several customized TOD benchmarks are proposed (\textit{e.g.} AI-TOD~\cite{AI-TOD_2020_ICPR}, TinyPerson~\cite{TinyPerson_2020_WACV}, and AI-TOD-v2~\cite{aitodv2_2022_isprs}), facilitating a series of downstream tasks including driving assistance, traffic management, and maritime rescue. Recently, TOD has gradually become a popular yet challenging direction independent of generic object detection~\cite{COCO_2014_ECCV,PASCAL_VOC_2015_IJCV}.

\begin{figure}[t]
\includegraphics[width=\linewidth]{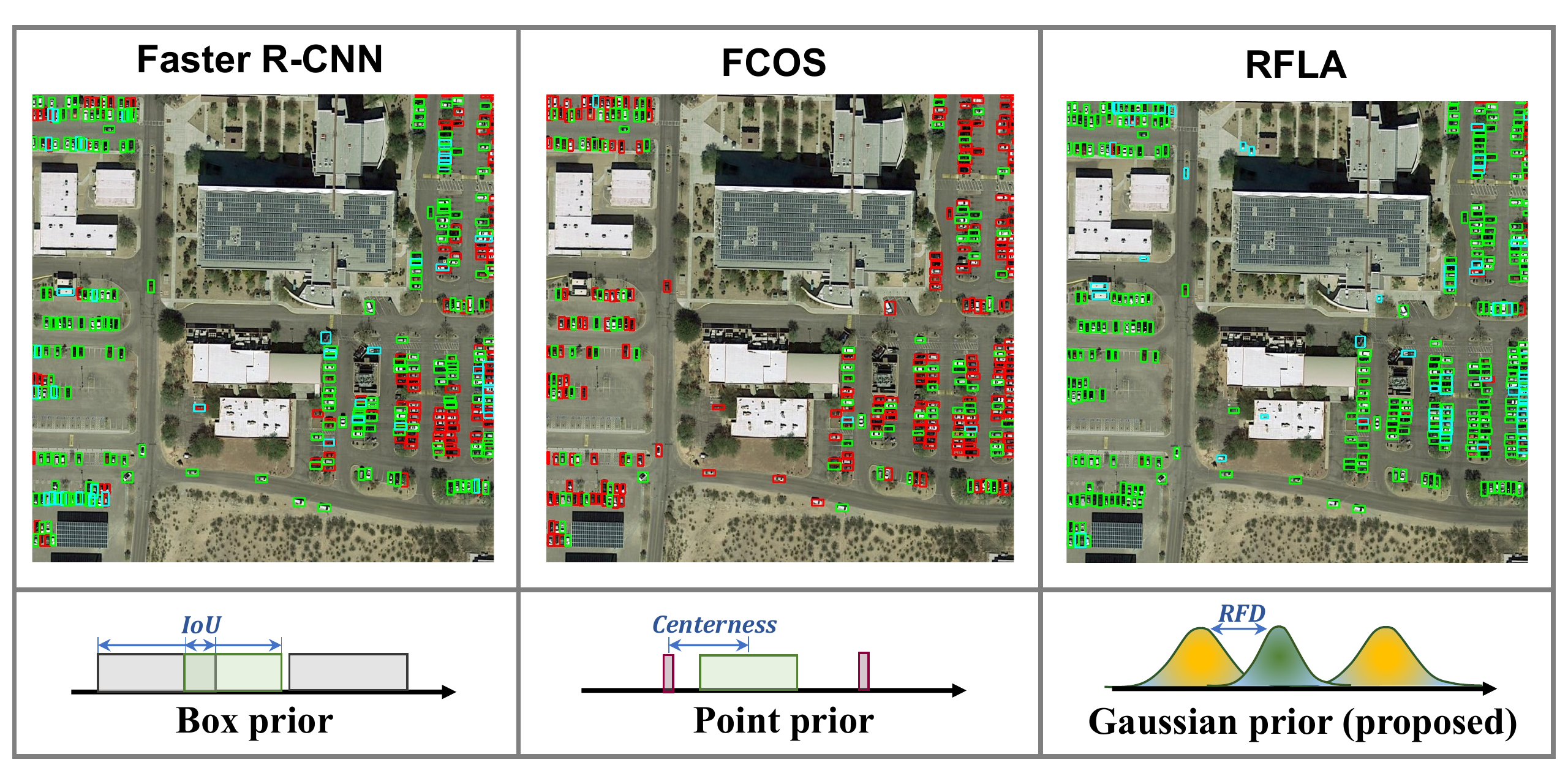}
\caption{Comparison between detection results of different label assignment schemes. The detection results are listed in the first row. The green, blue and red boxes denote true positive (TP), false positive (FP) and false negative (FN) predictions. The schematic diagram of different prior is in the second row, where the green region is \gt, grey, red and yellow regions denote box, point and Gaussian prior respectively.}
\label{fig:fig1}
\end{figure}

Generic object detectors can be divided into two factions: the anchor-based and the anchor-free paradigms. For anchor-based detectors, prior boxes of discrete locations, scales, and aspect ratios are heuristically preset. Then, label assignment strategies (\textit{e.g.} Max IoU Strategy~\cite{Faster-R-CNN_2015_NIPS}, ATSS~\cite{ATSS_CVPR_2020}) are constructed mainly based on IoU to find the appropriate matching relationship between anchors and ground truth (\gt). Anchor-free detectors change the prior from boxes to points. Usually the point prior covered by \gt \ is regarded as a positive sample (with the centerness in FCOS~\cite{FCOS_2019_ICCV}), saving the effort of anchor box fine-tuning. 

Despite the outstanding performance of the above two factions on generic object detection tasks, their performance on TOD tasks commonly suffers a drastic drop~\cite{AI-TOD_2020_ICPR,TinyPerson_2020_WACV}. In this paper, we argue that the current prior box and point along with their corresponding measurement strategies are sub-optimal for tiny objects, which will further hinder the process of label assignment. Specifically, we take the individual prior box and point as instances and rethink them from the perspective of distribution.
\begin{equation}
    p(v|x,y) = \frac{\varepsilon(x-x_{1})\varepsilon(x_{2}-x)\varepsilon(y-y_{1})\varepsilon(y_{2}-y)}{(x_{2}-x_{1})(y_{2}-y_{1})}
    \label{eq.1}
\end{equation}
where $p(v|x,y)$ is the probability density function of prior information, $(x, y)$ is the location on the image, $v$ is the weight of the corresponding location, $\varepsilon(\cdot)$ is a step function that equals to 1 when the input is larger than 0, otherwise equals to 0. $[(x_{1}, y_{1}), (x_{2}, y_{2})]$ is the region of prior information, for anchor-based detectors $x_{2}-x_{1}=\text{width},~y_{2}-y_{1}=\text{height}$, while for anchor-free detectors $x_{2}-x_{1}=1,~y_{2}-y_{1}=1$. The schematic diagram of different prior is shown in the second row of Fig.~\ref{fig:fig1}, existing prior information combing with its corresponding measurement strategy has the following problems for tiny objects. 

First, the individual box prior and point prior both have a limited prior domain (where $p(v|x,y)>0$), while existing label assignment metrics are highly dependent on the overlap of domain. In other words, when a particular \gt \ has no overlap with a specific prior, their positional relationship cannot be solved by IoU or centerness. 
For tiny objects, it is often the case that the \gt \ box has no overlap with almost all anchor boxes (\textit{i.e.} $\rm{IoU=0}$) or does not contain any anchor points~\cite{dotd_2021_cvprw}, leading to tiny objects' lack of positive samples~\cite{dotd_2021_cvprw}. 
To this end, heuristics are usually deployed to guarantee more positive samples for tiny objects~\cite{ATSS_CVPR_2020,s3fd_2017_cvpr}. However, the assigner often fails to compensate positive samples for tiny objects based on the zero-valued IoU or centerness. Therefore, the network will impose less attention on tiny object learning. Details about this point are analyzed in Sec.~\ref{sec.analysis}.
Second, current prior region mainly follows a uniform distribution and treats each location inside the prior region equally ($v=constant$). However, prior information is essentially leveraged to assist the label assignment or feature point assignment process~\cite{ATSS_CVPR_2020}. In this process, one implicit rule is assigning feature points with appropriate receptive field to \gt~\cite{Faster-R-CNN_2015_NIPS,FCOS_2019_ICCV}. As theoretically analyzed in previous work~\cite{erf_2016_nips}, when remapping the receptive field of feature point back onto the input image, the effective receptive field is actually Gaussian distributed. The gap between the uniformly distributed prior and the Gaussian distributed receptive field will lead to the mismatch between \gt \ and the receptive field of the feature points assigned~to~it.  

To mitigate the above problems, we introduce a novel prior based on Gaussian distribution and build a Gaussian Receptive Field based Label Assignment (RFLA) strategy that is more conducive to tiny objects. Specifically, we propose to directly measure the similarity between the Gaussian receptive field and \gt \ region with a newly designed Receptive Field Distance (RFD). 
Leveraging the Gaussian receptive field as prior information can elegantly address the issues incurred by box and point prior.
On the one hand, the Gaussian distribution is not step changed. The domain of each individual prior is the entire image, where the weight of each location gradually decays from the center to the periphery with a value higher than 0. It is thus feasible to model the positional relationship between any feature point and any \gt \ on the whole image, making it possible to obtain balanced positive samples for different sized objects. On the other hand, Gaussian prior can better fit the property of Gaussian effective receptive field, thereby alleviating receptive field mismatch problem, especially for tiny~objects.

Moreover, since IoU and RFD are not in the same dimension, directly applying the new metric to the existing threshold-based label assignment structure is not rational. Instead, we introduce to rank the priority of each feature point \textit{w.r.t.}~their RFD scores, based on which we further design a Hierarchical Label Assigner (HLA) which progressively alleviates outlier \gt~samples and obtain sufficient training for tiny objects. 

The contributions of this paper are summarized as follows:
\begin{enumerate}[(1)]
    \item We experimentally reveal that current anchor-based and anchor-free detectors exist scale-sample imbalance problem in tiny object label assignment.
    \item To mitigate the above problem, we introduce a simple but effective Receptive Field-based Label Assignment (RFLA) strategy. The RFLA is easy to replace the standard box and point-based label assignment strategies in mainstream detectors, boosting their performance on TOD.
    \item Extensive experiments on four datasets validate the performance superiority of our proposed method. The introduced method significantly outperforms the state-of-the-art competitors on the challenging AI-TOD dataset without additional costs in the inference stage.
\end{enumerate}

\section{Related Work}

\subsection{Object Detection}

The mainstream object detection methods include anchor-based detectors and anchor-free detectors. Classic anchor-based detectors include Faster R-CNN~\cite{Faster-R-CNN_2015_NIPS}, Cascade R-CNN~\cite{Cascade-R-CNN_2018_CVPR}, RetinaNet~\cite{Focal-Loss_2017_ICCV}, YOLO series~\cite{YOLOv2_2017_CVPR,YOLOv3_2017,Yolov4_2020_arXiv} \etc \ It is commonly believed that one fundamental defect of the anchor-based paradigm is its requirement of tuning \textit{w.r.t}~the specific task~\cite{FCOS_2019_ICCV}. Moreover, the IoU-based label assignment strategy~\cite{Faster-R-CNN_2015_NIPS} that is built upon anchor boxes also introduces additional hyper-parameters, showing a significant impact on the detection performance. 

Anchor-free detectors get rid of the constraints of anchor-boxes, and seek to directly predict objects from center points like FCOS~\cite{FCOS_2019_ICCV} and FoveaBox~\cite{FoveaBox_2020_TIP}, or seek to predict objects from key-points such as CornerNet~\cite{CornerNet_2018_ECCV}, Grid R-CNN~\cite{Grid_RCNN_2019_CVPR} and RepPoints~\cite{RepPoints_2019_ICCV}. The recently published anchor-free detectors mainly follow the end-to-end paradigm, they merely preset a set of boxes without shape or location prior information, and then directly reason about the final predictions, such as DETR~\cite{detr_2020_eccv}, Deformable DETR~\cite{deformabledetr_2021_iclr}, and Sparse R-CNN~\cite{sparsercnn_2021_cvpr}. Despite the success of the end-to-end paradigm on generic object detection tasks, their performance on TOD tasks require further investigation.

Unlike box and point prior-based detectors, we introduce another prior information based on the receptive field. Combining the Gaussian receptive field and its customized label assignment strategy can significantly alleviate the imbalance problem raised by existing prior and measurement for tiny objects.

\subsection{Tiny Object Detection}

Most of the existing tiny object detection methods can be roughly grouped into the following four classes: Data augmentation, Multi-scale learning, Customized training strategy for tiny objects, and Feature enhancement strategy.

\noindent \textbf{Data augmentation.}  A simple yet effective way is to collect more tiny object data. Another way is to use simple data augmentations include rotating, image flipping, and up-sampling. \ Krisantal \textit{et al.}~\cite{augmentation_2019_arxiv} seek to enhance TOD performance by oversampling images that contain tiny objects and copy-pasting~them.

\noindent \textbf{Multi-scale learning.} The multi-resolution image pyramid is a basic way of multi-scale learning. To reduce the computation cost, some works~\cite{SSD_2016_ECCV,FPN_2017_CVPR,M2Det_2019_AAAI} propose to construct feature-level pyramid. After that, lots of methods attempt to further improve FPN, some of them are PANet~\cite{PANet_2018_CVPR}, BiFPN~\cite{Efficientdet_2020_CVPR}, Recursive-FPN~\cite{DetectoRS_2020_CVPR}. Besides, TridentNet~\cite{Trident-Net_2019_ICCV} constructs multi-branch detection heads with different receptive fields to generate scale-specific feature maps. Multi-scale learning strategies commonly boost TOD performance with additional computation.

\noindent \textbf{Customized training strategy for tiny objects.} Object detectors usually cannot get satisfactory performance on tiny objects and large objects simultaneously. Inspired by this fact, SNIP~\cite{SNIP_2018_CVPR} and SNIPER~\cite{SNIPER_2018_NIPS} are designed to selectively train objects within a certain scale range. In addition, Kim~\textit{et al.}~\cite{san_2018_eccv} introduces a Scale-Aware Network (SAN) and maps the features of different spaces onto a scale-invariant subspace, making detectors more robust to scale variation.

\noindent \textbf{Feature enhancement strategy.} Some works propose to enhance the feature representation of small objects by super-solution or GAN. PGAN~\cite{PGAN_2017_CVPR} makes the first attempt to apply GAN to small object detection. Moreover, Bai~\textit{et al.}~\cite{SOD-MTGAN_2018_ECCV}~propose an MT-GAN which trains an image-level super-resolution model for enhancing the small RoI features. Feature-level super-resolution~\cite{Better_to_Follow_2019_ICCV} is proposed to improve small object detection performance for proposal based detectors. Also, there are some other super-solution based methods including ~\cite{auxiliarygan_2021_rs,residualsuperres_2021_rs,edgegan_2020_rs}.

Most of the methods dedicated to TOD will bring about additional annotation or computation costs. In contrast, our proposed method attempts to push forward TOD from the perspective of label assignment, and our proposed strategy will not bring any additional cost in the inference stage.

\subsection{Label Assignment in Object Detection}

As revealed by ATSS~\cite{ATSS_CVPR_2020}, the essential difference between the anchor-free and anchor-based detector is the way of defining training samples. The selection of positive and negative (\posneg) training samples will notably affect the detector's performance. Recently, many works have been proposed for better label assignment in generic object detection tasks. FreeAnchor~\cite{Freeanchor_2021_TPAMI} decides positive anchors based on a detection-customized likelihood. PAA~\cite{PAA_2020_ECCV} proposes to use GMM to model the distribution of anchors and divide \posneg~samples based on the center of GMM. OTA~\cite{ota_2021_cvpr} models the label assignment process as an optimal transport problem and seeks to solve the optimal assignment strategy. ATSS~\cite{ATSS_CVPR_2020} adaptively  adjusting the \posneg~samples \textit{w.r.t.}~their statistics characteristics. AutoAssign~\cite{autoassign_2020_arxiv} and IQDet~\cite{iqdet_2021_cvpr} reweight and sample high-quality regions based on the predicted IoU and confidence.  

Unlike the above-mentioned general object detection strategies, this paper focuses on the design of prior information and its corresponding label assignment strategy for TOD.

\section{Method}
\subsection{Receptive Field Modelling}

One basic principle that mainstream object detectors obey is \textit{dividing and conquering}, namely detecting objects of different scales on the different layers of FPN~\cite{yolof_2021_cvpr,FPN_2017_CVPR}.
Specifically, anchor-based detectors tile prior boxes of different scales on different layers of FPN to assist label assignment, and objects of different scales are thus detected on the different layers of FPN.  For anchor-free detectors, they group objects in different scale ranges (\textit{e.g.}~[0, 64] for $P_{3}$) onto different levels of FPN for detection. Despite label assignment strategy varies, one common ground of anchor-based and anchor-free detectors is to assign feature points of an appropriate receptive field to objects of different scales~\cite{Faster-R-CNN_2015_NIPS,FCOS_2019_ICCV}. Thus, the receptive field can directly serve as a founded and convincing prior for label assignment without the designing of heuristic anchor box preset or scale grouping.

In this paper, we propose to directly measure the matching degree between the Effective Receptive Field (ERF) and the \textit{gt} region for label assignment, getting rid of the box or point prior that deteriorates TOD. Previous work has pointed out that the ERF can be theoretically derived as Gaussian distribution~\cite{erf_2016_nips}. In this work, we follow this paradigm and seek to model the ERF of each feature point as Gaussian distribution, and we first derive the Theoretical Receptive Field (TRF) of the $n\mbox{-}th$ layer on a standard convolution neural network~\cite{ResNet_2016_CVPR} by the following formula as $tr_{n}$:
\begin{equation}
    \setlength{\abovedisplayskip}{\equspace}
	\setlength{\belowdisplayskip}{\equspace}
    tr_{n}=tr_{n-1}+\left(k_{n}-1\right) \prod_{i=1}^{n-1} s_{i}
\end{equation}
where $tr_{n}$ denotes the TRF of each point on the $n\mbox{-}th$ convolution layer, $k_{n}$ and $s_{n}$ denotes the kernel size and stride of the convolution operation on the $n\mbox{-}th$~layer. 

As studied in~\cite{erf_2016_nips}, the ERF and TRF have the same center points but the ERF of each feature point only occupies part of the full TRF. Therefore, we use the location of each feature point $(x_{n}, y_{n})$ as the mean vector of a standard 2-D Gaussian distribution. As it is hard to get the precise ERF, we approximate the ERF radius $er_{n}$ with half the radius of TRF. The square of $er_{n}$ serves as the co-variance of 2-D Gaussian distribution for a standard square-like convolution kernel. To sum up, we model the range of ERF into a 2-D Gaussian distribution $N_{e}(\boldsymbol{\mu_{e}}, \boldsymbol{\Sigma_{e}})$ with
\begin{equation}
    \setlength{\abovedisplayskip}{\equspace}
	\setlength{\belowdisplayskip}{\equspace}
    \boldsymbol{\mu_{e}}=\begin{bmatrix}
        x_{n}\\ y_n
    \end{bmatrix}
    ,
    \mathbf{\Sigma_{e} }=\begin{bmatrix}
    er^{2}_{n} & 0 \\ 
    0 & er^{2}_{n}
    \end{bmatrix}.
    \label{eq3.3}
\end{equation}

\begin{figure}[t]
\includegraphics[width=\linewidth]{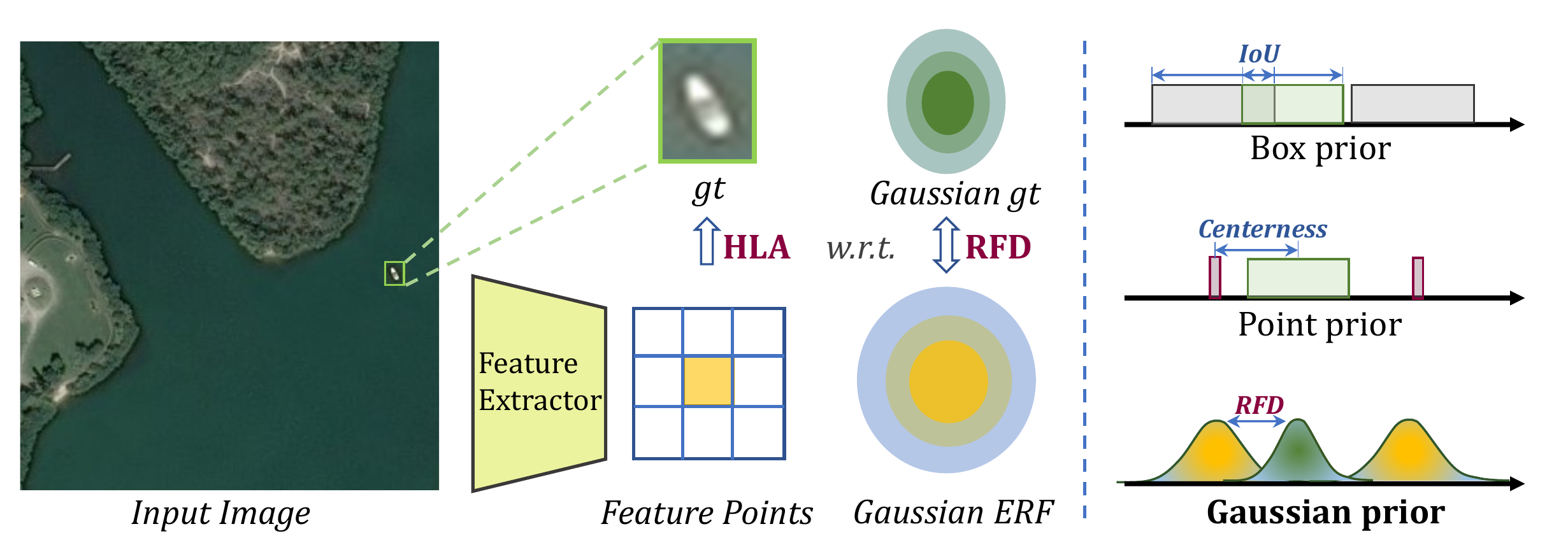}
\caption{The process of RFLA. In RFLA, we directly calculate the RFD between Gaussian ERF and \gt, then assign labels with HLA.}
\label{fig:rfla}
\end{figure}

\subsection{Receptive Field Distance}
\label{sec.rfd}
Obtaining the Gaussian ERF, the following key step is to measure the matching degree between the ERF of feature points and a certain \textit{gt}. As discussed in the introduction, the step-changed uniform distribution is not conducive to tiny objects, it is also necessary to model \gt \ into another distribution.

Observing that the main body of object is aggregated in the center of bounding box~\cite{CenterMap-Net_2020_TGRS,nwd_2021_arxiv}, we also model the \textit{gt} box $(x_{g}, y_{g}, w_{g}, h_{g})$ into a standard 2-D Gaussian distribution $N_{g}(\boldsymbol{\mu_{g}}, \boldsymbol{\Sigma_{g}})$, where the center point of each annotated box serves as the mean vector of Gaussian and the square of half side length serves as the co-variance matrix, namely, 
\begin{equation}
    \setlength{\abovedisplayskip}{\equspace}
	\setlength{\belowdisplayskip}{\equspace}
    \boldsymbol{\mu_{g}}=\begin{bmatrix}
        x_{g}\\ y_g
    \end{bmatrix}
    ,
    \boldsymbol{\Sigma_{g}}=\begin{bmatrix}
    \frac{w_{g}^{2}}{4} & 0 \\ 
    0 & \frac{h_{g}^{2}}{4}
    \end{bmatrix}.
    \label{eq3.3}
\end{equation}

In this paper, we investigate three types of classic distances between Gaussian distributions as Receptive Field Distance Candidates (RFDC). These distance measurement include Wasserstein distance~\cite{optimaltransport_2019,gwd_2021_icml}, K-L divergence~\cite{KL_divergence_Book_2007,kld_2021_nips} and J-S divergence~\cite{JS_divergence_TIT_2003}. The J-S divergence between Gaussian distributions has no closed-form solution~\cite{JS_divergence_TIT_2003,gjsd_2019_entropy}, enormous computation will be introduced when approximating its solution, thus, the J-S divergence is not used. Herein, we will first analyze their closed form solutions in our task, then discuss their pros and cons for the TOD task. 

\textbf{Wasserstein distance.} The Wasserstein Distance comes from Optimal Transport theory~\cite{optimaltransport_2019}. 
Given the Gaussian ERF $n_{e}=N_{e}(\boldsymbol{\mu_{e}}, \boldsymbol{\Sigma_{e}})$ and Gaussian \textit{gt} $n_{g}=N_{g}(\boldsymbol{\mu_{g}}, \boldsymbol{\Sigma_{g}})$, the $2^{nd}$ Wasserstein distance can be simplized as Eq.~\ref{eq.gwd}~\cite{gwd_2021_icml}.
\begin{equation}
    \setlength{\abovedisplayskip}{\equspace}
	\setlength{\belowdisplayskip}{\equspace}
    W_{2}^{2}(n_{e}, n_{g}) = \left\|\left(\left[x_{n}, y_{n}, er_{n}, er_{n}\right]^{\mathrm{T}},\left[x_{g}, y_{g}, \frac{w_{g}}{2}, \frac{h_{g}}{2}\right]^{\mathrm{T}}\right)\right\|_{2}^{2}.
    \label{eq.gwd}
\end{equation}

The main advantage of Wasserstein Distance is that it can measure two non-overlapping distributions~\cite{optimaltransport_2019}. It is always the case that the \textit{gt} box has no overlap with most prior box and points, and the assigner fails to rank the priority of these candidates to a certain \textit{gt}. Thus, it is easy to say the characteristic of Wasserstein distance is conducive to TOD, which can consistently reflect the matching degree between all feature points and a certain \textit{gt} box, making the assigner feasible to compensate more positive samples for tiny objects according to a rational priority. However, the Wasserstein distance is not scale invariant and might be sub-optimal when the dataset contains objects of large-scale variance~\cite{kld_2021_nips}.

\textbf{Kullback-Leibler divergence.} Kullback-Leibler Divergence (KLD) is a classic statistical distance which measures how one probability distribution is different from another. 
KLD between two Gaussian distributions also has a closed form solution, the KLD between ERF $n_{e}$ and \textit{gt} region $n_{g}$ is as follows:
\begin{equation}
    \setlength{\abovedisplayskip}{\equspace}
	\setlength{\belowdisplayskip}{\equspace}
    D_{\mathrm{KL}}\left(n_{e} \| n_{g}\right)=\frac{1}{2}( \operatorname{tr}\left(\boldsymbol{\Sigma}_{g}^{-1} \boldsymbol{\Sigma}_{e}\right)+ \left(\boldsymbol{\mu}_{g}-\boldsymbol{\mu}_{e}\right)^{\top} \boldsymbol{\Sigma}_{g}^{-1}\left(\boldsymbol{\mu}_{g}-\boldsymbol{\mu}_{e}\right)+\ln \frac{\left|\boldsymbol{\Sigma}_{g}\right|}{\left|\boldsymbol{\Sigma}_{e}\right|}-2),
    \label{gaussiankld}
\end{equation}
Eq.~\ref{gaussiankld} can be further simplified as:

\begin{equation}
    \setlength{\abovedisplayskip}{\equspace}
	\setlength{\belowdisplayskip}{\equspace}
     D_{\mathrm{KL}}\left(n_{e} \| n_{g}\right)=\frac{2er_{n}^{2}}{w_{g}^{2}}+\frac{2er_{n}^{2}}{h_{g}^{2}}+\frac{2 (x_{n}-x_{g})^{2}}{w_{g}^{2}}+\frac{2(y_{n}-y_{g})^2}{h_{g}^{2}}+\ln \frac{w_{g}}{2er_{n}}+\ln \frac{h_{g}}{2er_{n}}-1.
\end{equation}

As demonstrated by the work~\cite{kld_2021_nips}, KLD has the property of scale invariance between two 2-D Gaussian distributions, and the scale invariance is crucial for the detection~\cite{Unitbox_2016_ACMM}. While the main disadvantage of KLD is that it cannot consistently reflect the distance between two distributions when their overlap is negligible. Hence, the KLD between ERF and \gt \ is selected as another RFDC in this paper. 

In summary, we investigate three classic ways of probability distribution measurement, while Wasserstein distance and KLD are selected as RFDC. Then, we apply a non-linear transformation into RFDC and get the RFD with a normalized value range between (0, 1) as follows:
\begin{equation}
    \setlength{\abovedisplayskip}{\equspace}
	\setlength{\belowdisplayskip}{\equspace}
    \rm{RFD} = \frac{1}{1+\rm{RFDC}}
\end{equation}

\subsection{Hierarchical Label Assignment}
\label{sec.hla}
Some anchor-based detectors set a threshold based on IoU to decide \posneg \ samples~\cite{Faster-R-CNN_2015_NIPS,Focal-Loss_2017_ICCV,Cascade-R-CNN_2018_CVPR}, while anchor-free detectors mainly divide \posneg \ samples by the spatial location between point prior and \gt \ region. Since tiny objects are usually unwelcome in both threshold-based and \gt \ region-based strategies, we propose to hierarchically assign labels to tiny objects by score ranking.

To guarantee that the positional relationship between any feature point and any \gt \ can be solved, the proposed Hierarchical Label Assignment (HLA) strategy is built on the proposed RFD. Before assigning, an RFD score matrix between feature points and \textit{gt} is computed based on the above method. In the first stage, we rank each feature point to its RFD score with a certain \textit{gt}. Then, positive labels are assigned to feature points with top k RFD scores with a certain \textit{gt}. Finally, we get the assigning result $r_{1}$ and the corresponding mask $m$ of features that have been assigned, where $m$ is binary-valued (0/1). In the second stage, to improve the overall recall and alleviate outliers, we slight decay the effective radius $er_{n}$ by multiplying a stage factor $\beta$, then repeat the above ranking strategy and supplement one positive sample to each \textit{gt}, getting the assigning result $r_{2}$. We obtain the final assigning result $r$ by the following rule:
\begin{equation}
    \setlength{\abovedisplayskip}{\equspace}
	\setlength{\belowdisplayskip}{\equspace}
    r = r_{1} m + r_{2}(1- m),
\end{equation}where the mask operation $m$ is taken to avoid introducing too many low-quality samples for those \textit{gt} which have already been assigned with sufficient samples. Not that the occluded sample will be assigned to the smaller \textit{gt}. Combining the RFD with the HLA strategy, we can get the complete Receptive Field based Label Assignment (RFLA) strategy for TOD.

\subsection{Application to Detectors}

The proposed RFLA strategy can easily be applied to anchor-based and anchor-free frameworks.  Without losing generality, we take the classic Faster R-CNN~\cite{Faster-R-CNN_2015_NIPS} and FCOS~\cite{FCOS_2019_ICCV} as examples. Concretely, for Faster R-CNN, RFLA can be used to replace the standard anchor tiling and MaxIoU anchor assigning process. For FCOS, we remove the constraint of limiting feature points inside \gt \ box because the tiny box only covers an extremely limited region which commonly holds much fewer feature points than the large object. Then, it is easy to replace the point based assigning with RFLA for balanced learning. Note that we modify the centerness~\cite{FCOS_2019_ICCV} loss into the following formula to avoid gradient explosion:
\begin{equation}
    \setlength{\abovedisplayskip}{\equspace}
	\setlength{\belowdisplayskip}{\equspace}
    \text {centerness}^{*}=\sqrt{\frac{\varepsilon[\min \left(l^{*}, r^{*}\right)]\min \left(l^{*}, r^{*}\right)+c}{\max \left(l^{*}, r^{*}\right)} \times \frac{\varepsilon[\min \left(t^{*}, b^{*}\right)]\min \left(t^{*}, b^{*}\right)+c}{\max \left(t^{*}, b^{*}\right)}},
\end{equation}
where $l^{*}, t^{*}, r^{*}, b^{*}$ are regression targets defined in FCOS, $\varepsilon(\cdot)$ is a step function same as that in Eq.~\ref{eq.1}, $c$ is a factor set to 0.01 to avoid gradient vanishing problem when the center point of regression target is outside the \gt \ box.
In the following part, extensive experiments will show RFLA's outstanding robustness to TOD.
\section{Experiment}
\subsection{Dataset}
Experiments are conducted on four datasets. The main experiments are performed on the challenging AI-TOD~\cite{AI-TOD_2020_ICPR} dataset, which has the smallest average absolute object size of 12.8 pixels and contains 28,036 images. Furthermore, we test the proposed method on TinyPerson~\cite{TinyPerson_2020_WACV}, VisDrone2019~\cite{visdrone2019_2019_iccvw} and DOTA-v2.0~\cite{DOTA2.0_2021_pami}. Note that the selected datasets all contain a great amount of tiny objects (smaller than $16\times16$ pixels).

\subsection{Experiment settings}
All the experiments are conducted on a computer with 1 NVIDIA RTX 3090 GPU, and the model training is based on PyTorch~\cite{PyTorch_2019_NIPS}, the core codes are built upon MMdetection~\cite{mmdetection_2019_arXiv}. The ImageNet~\cite{ImageNet_2015_IJCV}
pre-trained model is used as the backbone. All models are trained using the Stochastic Gradient Descent (SGD) optimizer for 12 epochs with 0.9 momenta, 0.0001 weight decay, and 2 batch size. The initial learning rate is set to 0.005 and decays at the $8^{th}$ and $11^{th}$ epochs. Besides, the number of RPN proposals is set to 3000. In the inference stage, the confidence score is set to 0.05 to filter out background bounding boxes,
and the NMS IoU threshold is set to 0.5 with top 3000 bounding boxes. All the other parameters are set the same as default in MMdetection. The evaluation metric follows AI-TOD benchmark~\cite{AI-TOD_2020_ICPR} except for experiments on TinyPerson. The above parameters are used in all experiments unless specified otherwise.
\subsection{Ablation study}
\label{sec.ablation_study}

\begin{table}[t]\scriptsize
\centering
\begin{minipage}[]{0.48\textwidth}
    \centering
    \caption{Comparison between different ways of measuring receptive-field distance. Note that in this experiment, all distances are built on HLA. }
    \begin{tabular}{l|cc|cc}  
	\toprule
	Distance  & AP & $\rm{AP_{0.5}}$ &  $\rm{AP_{vt}}$ & $\rm{AP_{t}}$  \\
	\midrule
	GIoU     &  17.9 & 45.1 & 5.5 & 16.7  \\
	WD     &  \textbf{21.1}	& \textbf{52.2}	& 6.6 & \textbf{21.5}    \\ 
	KLD   &  \textbf{21.1} &  51.6 & \textbf{9.5}  & 21.2\\
	\bottomrule
    \end{tabular}
    \label{distance_metrics}
\end{minipage}
\begin{minipage}[]{0.48\textwidth}
    \centering
    \caption{Influence of different designs. Note that RFD denotes only using the first stage of HLA, HLA means using all stages of HLA. }
    \begin{tabular}{cc|cc|cc}  
	\toprule
	  RFD & HLA &  AP & $\rm{AP_{0.5}}$  & $\rm{AP_{vt}}$ & $\rm{AP_{t}}$  \\
	\midrule
	     &  	&	11.1	& 26.3	 & 0.0 & 7.2    \\
	  \checkmark    & &  20.7	&	50.6 &  7.6  &  20.5       \\
	  \checkmark  & \checkmark & \textbf{21.1} & \textbf{51.6} & \textbf{9.5} & \textbf{21.2}   \\
	\bottomrule
    \end{tabular}
    \label{individual}
\end{minipage}
\end{table}

\begin{table}[t]\scriptsize
\centering
\begin{minipage}[]{0.48\textwidth}
    \centering
    \caption{Influence of the stage factor $\beta$ in the Hierarchical Label Assigner (HLA). The KLD is used as RFD.} 
    \begin{tabular}{c|cc|cc}  
	\toprule
	 $\beta$ & AP & $\rm{AP_{0.5}}$ & $\rm{AP_{vt}}$ & $\rm{AP_{t}}$   \\
	\midrule
	0.95	& \textbf{21.1}	& 51.1 & 8.4 & 21.1     \\
	0.9  	& \textbf{21.1}	& \textbf{51.6}  & \textbf{9.5} & \textbf{21.2}   \\
	0.85	& 20.8	& 51.4  & 7.4 & 21.0   \\
	0.8		& 19.7	& 49.0  & 5.6 & 19.2  \\
	\bottomrule
    \end{tabular}
    \label{decay_factor}
\end{minipage}
\begin{minipage}[]{0.48\textwidth}
    \centering
    \caption{Gaussian anchor and receptive anchor. GA means Gaussian Anchor and RA means Receptive Anchor.}
    \begin{tabular}{c|cc|cc}  
	\toprule
	Method  & AP & $\rm{AP_{0.5}}$  & $\rm{AP_{vt}}$ & $\rm{AP_{t}}$  \\
	\midrule
	GA	&  19.6	& 49.2  & 8.2 &  19.7 \\
	RA   & 18.9	& 47.5 & 6.1 & 19.1   \\
	baseline & 11.1 & 26.3 & 0.0 & 7.2 \\
	\bottomrule
    \end{tabular}
    \label{gaussian_anchor}
\end{minipage}
\end{table}

\setlength{\tabcolsep}{4pt}
\begin{table}[t]
\caption{Main results on AI-TOD. Note that models are trained on the {\tt trainval set} and validated on the {\tt test set}. Note that FCOS* means using P2-P6 of FPN.}
\label{table:main_resuls}
\renewcommand{\arraystretch}{0.85}
\renewcommand{\tabcolsep}{1.0mm}
\resizebox{\linewidth}{!}{
\begin{tabular}{l|c|ccc|ccccc}
	\toprule
	Method  & Backbone & AP & $\rm{AP_{0.5}}$ & $\rm{AP_{0.75}}$ & $\rm{AP_{vt}}$ & $\rm{AP_{t}}$ & $\rm{AP_{s}}$ & $\rm{AP_{m}}$  \\
	\midrule
	  TridentNet~\cite{Trident-Net_2019_ICCV}      & ResNet-50 	 & 7.5	& 20.9 & 3.6 & 1.0 & 5.8 & 12.6 & 14.0    \\
	  Faster R-CNN~\cite{Faster-R-CNN_2015_NIPS}   & ResNet-50 	 	& 11.1	& 26.3 & 7.6 & 0.0 & 7.2 & 23.3 & 33.6  \\
	  Cascade RPN~\cite{cascade_rpn_2019_nips} & ResNet-50	 	& 13.3	& 33.5  & 7.8 & 3.9 & 12.9 & 18.1 & 26.3 \\
	  Cascade R-CNN~\cite{Cascade-R-CNN_2018_CVPR}  & ResNet-50 	 & 13.8	& 30.8 & 10.5 & 0.0 & 10.6 & 25.5 & 36.6 \\
	  DetectoRS~\cite{DetectoRS_2020_CVPR}  & ResNet-50 	 	& 14.8	& 32.8 & 11.4 & 0.0 & 10.8 & 28.3 & 38.0  \\
	  DotD~\cite{dotd_2021_cvprw} & ResNet-50 & 	16.1 & 39.2  & 10.6 & 8.3 & 17.6 & 18.1 & 22.1     \\
	  DetectoRS w/ NWD~\cite{nwd_2021_arxiv} & ResNet-50 & 	20.8 & 49.3  & 14.3 & 6.4 & 19.7 & 29.6 & \textbf{38.3}    \\
    \midrule
    SSD-512~\cite{SSD_2016_ECCV}        	   & ResNet-50 	 & 7.0	& 21.7 & 2.8 & 1.0 & 4.7 & 11.5 & 13.5  \\
    RetinaNet~\cite{Focal-Loss_2017_ICCV}      & ResNet-50	 & 8.7	& 22.3 & 4.8 & 2.4 & 8.9 & 12.2 & 16.0     \\
    PAA~\cite{PAA_2020_ECCV} & ResNet-50 	 & 10.0	& 26.5 & 6.7 & 3.5 & 10.5 & 13.1 & 22.1 \\
	 ATSS~\cite{ATSS_CVPR_2020}   & ResNet-50 	 & 12.8	& 30.6 & 8.5 & 1.9 & 11.6 & 19.5 & 29.2     \\
	\midrule
	 RepPonits~\cite{RepPoints_2019_ICCV}       & ResNet-50	 	& 9.2	& 23.6 & 5.3 & 2.5 & 9.2 & 12.9 & 14.4   \\
	 OTA~\cite{ota_2021_cvpr}  & ResNet-50 	& 10.4	& 24.3  & 7.2 &  2.5 & 11.9 & 15.7 & 20.9   \\
	 AutoAssign~\cite{autoassign_2020_arxiv}  & ResNet-50 	& 12.2	& 32.0  & 6.8 &  3.4 & 13.7 & 16.0 & 19.1   \\
     FCOS~\cite{FCOS_2019_ICCV}                 & ResNet-50 	& 12.6	& 30.4 & 8.1 & 2.3 & 12.2 & 17.2 &  25.0 \\
     M-CenterNet~\cite{AI-TOD_2020_ICPR} & DLA-34 & 14.5 & 40.7 & 6.4 & 6.1 & 15.0 & 19.4 & 20.4 \\ 
     FCOS*                & ResNet-50 	& 15.4	& 36.3 & 10.9 & 6.0 & 17.6 & 18.5 &  20.7 \\

	\midrule
	 RetinaNet w/ RFLA    & ResNet-50 	  & 9.1	& 23.1 & 5.2 & 4.1 & 10.5 & 10.5 & 12.3 \\
	 AutoAssign w/ RFLA  & ResNet-50 	& 14.2	& 37.8  & 6.9 &  6.4 & 14.9 & 17.4 & 21.8   \\
	 FCOS* w/ RFLA &  ResNet-50 	 & 16.3	& 39.1 & 11.3 & 7.3 & 18.5 & 19.8 & 21.8    \\
	 Faster R-CNN w/ RFLA  & ResNet-50 	 	& 21.1	& 51.6 & 13.1 & \textbf{9.5} & 21.2 & 26.1 & 31.5  \\
	 Cascade R-CNN w/ RFLA  & ResNet-50  &	 22.1	& 51.6 & 15.6 & 8.2 & 22.0 & 27.3 & 35.2    \\
	 DetectoRS w/ RFLA & ResNet-50 	 & \textbf{24.8}	& \textbf{55.2} & \textbf{18.5} & 9.3 & \textbf{24.8} & \textbf{30.3} & 38.2 \\
	\bottomrule
	\end{tabular}}
\end{table}
\setlength{\tabcolsep}{1.4pt}

\noindent \textbf{Effectiveness of different RFD.} In this part, we respectively apply Wasserstein distance (WD) and Kullback-Leibler divergence (KLD) to measure the distance between Gaussian ERF and \textit{gt} region as discussed in Sec.~\ref{sec.rfd}. We also test the performance of GIoU~\cite{GIoU_loss_2019_CVPR} by setting the prior as ERF sized box. Note that all experiments are conducted based on Faster R-CNN w/ HLA since RFD and HLA are interdependent. As shown in Tab.~\ref{distance_metrics}, it can be seen that GIoU is inferior to RFD since it fails to distinguish the locations of mutually inclusive boxes, while the performance of WD and KLD are comparable. The KLD surpasses WD in $\rm{AP_{vt}}$, while slightly lower than WD under $\rm{AP_{t}}$ metric. As mentioned in Sec.~\ref{sec.rfd}, the KLD is scale-invariant, thus it is better for \textit{very tiny} objects. Note that in the following experiments, we use KLD as the default RFD.

\noindent \textbf{Effectiveness of individual component.} The core designs in this paper are interdependent, whilst they can be separated into two parts: the Hierarchical Label Assignment (HLA) strategy and the Receptive Field Distance (RFD) built upon the HLA. Note that the validation of RFD requires using the first stage of HLA, we do not assign labels based on the threshold of RFD since the original threshold in the baseline detector is designed for IoU, which is not in the same dimension as RFD. We progressively apply RFD and HLA into the Faster R-CNN. Results are listed in Tab.~\ref{individual}, AP improves progressively, the individual effectiveness is thus verified. When switching the IoU-based assignment strategy to the RFD-based one, a notable improvement of $9.6$ AP points is obtained. This can be explained that the limited domain of box prior leads to the remarkably low IoU between anchors and \gt, many \gt \ fail to match with any anchor. With Gaussian prior and RFD, the assigner is capable of measuring the priority (RFD score) of all feature points to a particular \gt. Thus, even though the \gt \ has no overlap with any box prior, some positive samples can be compensated for the \gt \ with a rational receptive field, leading to the sufficient training of tiny objects.  

\noindent \textbf{Performance of different decay factor $\beta$.} As in Sec.~\ref{sec.hla}, in the HLA, we design a stage factor $\beta$ to the ERF for mitigating the outlier effect. In Tab.~\ref{decay_factor}, we keep all other parameters fixed and experimentally show that 0.9 is the best choice. Setting $\beta$ to a lower value will introduce too many low-quality samples.

\begin{figure}[t]
\centering
\begin{minipage}[]{0.47\textwidth}
    \centering
    \includegraphics[width=0.7\textwidth, height=80pt]{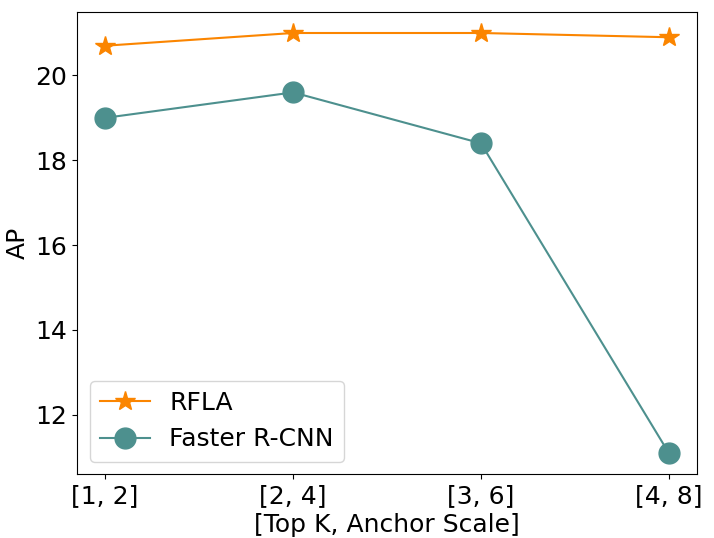}
    \caption{Comparison between top $k$ in Faster R-CNN w/ HLA and anchor fine-tuning in Faster R-CNN baseline.}
    \label{fig:anchor_scale}
   \end{minipage}
    \begin{minipage}[]{0.47\textwidth}
    \centering
    \includegraphics[width=0.7\textwidth, height=80pt]{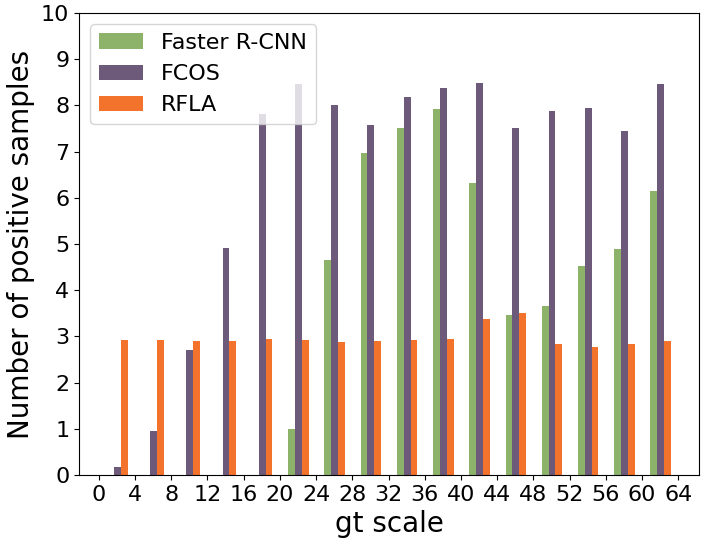}
    \caption{Scale-sample imbalance problems of different detectors. Base anchor scale is set to 8 for Faster R-CNN.}
    \label{fig:imbalance}
   \end{minipage}
   \vspace{-2mm}
\end{figure}

\noindent \textbf{Performance of different $k$.} In the HLA, the hyper-parameter $k$ is designed to adjust the number of positive samples assigned to each instance. Herein, we keep all other parameters fixed and set $k$ from $1$ to $4$. Their performance is $20.7$, $21.1$, $21.1$, and $20.9$ AP, respectively. When setting $k$ to $2$ or $3$, the best performance can be attained. Thus $3$ is recommended as the default setting. Moreover, the AP only waves by a small margin under the tested $k$. Moreover, we compare the AP under different $k$ with the result of anchor size tuning, as shown in Fig.~\ref{fig:anchor_scale}. It is easy to find that the performance of box prior based detector is quite sensitive to the box size on TOD tasks, while in our design, the performance is quite robust to the chosen of $k$, which consistently keeps a high level over box prior.

\noindent \textbf{Gaussian anchor and receptive anchor.} We directly model the anchors into Gaussian distributions, calculate the RFD score between \gt, and then assign labels with HLA. Results are shown in Tab.~\ref{gaussian_anchor}. The results show a great advantage of Gaussian prior and its combination with HLA. The Gaussian prior has a broader domain, making sample compensation possible. In addition, we change the anchor scale to the ERF scale, then assign labels with the MaxIoU strategy. The improvement over baseline further indicates the sensitivity of the box prior to detection performance for TOD.  It also reveals that the current anchor will potentially introduce the receptive field mismatch problem for tiny objects.

\subsection{Main result}
We compare our method with other state-of-the-art detectors on AI-TOD benchmark~\cite{AI-TOD_2020_ICPR}. 
As shown in Tab~\ref{table:main_resuls}, DetectoRS w/ RFLA achieves $24.8$ AP, which has $4.0$ AP above the state-of-the-art competitors. Notably, the improvement of RFLA with multi-stage anchor-based detectors is particularly significant. We think it mainly owes to the multi-stage detectors' mechanism of \textit{looking and thinking twice}. In the first stage, the combination of the proposed RFLA with RPN can improve the recall of tiny objects to a great extent. In the second stage, proposals are refined for precise location and classification. Besides, improvements can also be expected on one-stage anchor-based or anchor-free detectors, and the improvement in $\rm{AP_{vt}}$ is more obvious, $1.7$ points for RetinaNet and $1.3$ points for FCOS*. The gap between one-stage and multi-stage detectors is common for TOD~\cite{TinyPerson_2020_WACV,VisDrone-2018-Det_2018_ECCV,visdrone2019_2019_iccvw}. It mainly results from the lack of multi-stage regression, which is crucial for TOD.

\setlength{\tabcolsep}{4pt}
\begin{table}[t]
\scriptsize
\begin{center}
\caption{Results on TinyPerson.}
\label{table:tinyperson}
\resizebox{\linewidth}{!}{
\begin{tabular}{l|cccccccc}  
	\toprule
	Method  & $\rm{AP_{50}^{tiny}}$ & $\rm{AP_{50}^{tiny1}}$ & $\rm{AP_{50}^{tiny2}}$ & $\rm{AP_{50}^{tiny3}}$ & $\rm{AP_{50}^{small}}$ & $\rm{AP_{25}^{tiny}}$ & $\rm{AP_{75}^{tiny}}$  \\
	\midrule
	 FCOS  &  	 	23.4	& 9.8	& 22.7 & 34.7 & 39.2 & 43.8 & 1.7   \\
	  Faster R-CNN  &  	 	48.7	& 32.3	& 54.5 & 58.8 & 64.7 & 68.9 & \textbf{6.0}    \\
	\midrule
	 FCOS w/ RFLA  & 	 	$26.5^{+3.1}$	& $10.0^{+0.2}$	& $24.4^{+1.7}$ & $40.6^{+5.9}$ & $50.5^{+11.3}$ & $50.0^{+6.2}$ & $2.9^{+1.2}$  \\
	 Faster R-CNN w/ RFLA  & 	$\textbf{50.1}^{+1.4}$ 	& $\textbf{32.8}^{+0.5}$	& $\textbf{55.6}^{+1.1}$ & $\textbf{60.6}^{+1.8}$ & $\textbf{65.3}^{+0.6}$ & $\textbf{69.9}^{+1.0}$ & $5.9^{-0.1}$   \\
	\bottomrule
	\end{tabular}}
\end{center}
\end{table}
\setlength{\tabcolsep}{1.4pt}

\begin{table}[t]\scriptsize
\centering
\begin{minipage}[]{0.48\textwidth}
    \centering
    \caption{Results on VisDrone2019. The {\tt train}, {\tt val} sets are used for training and validation. FR, DR denote Faster R-CNN, DetectoRS, * means \textit{with RFLA}.  }
    \begin{tabular}{l|cc|cc}  
	\toprule
	Method  & AP & $\rm{AP_{0.5}}$ &  $\rm{AP_{vt}}$ & $\rm{AP_{t}}$  \\
	\midrule
	FCOS     &  14.1 & 25.5 & 0.1 & 2.1  \\
	FR    &  22.3	& 38.0	& 0.1& 6.2    \\
	DR   &  25.7 &  41.7 & 0.5  & 7.6\\
	\midrule
	FCOS*    & 	 	$15.1^{+1.0}$	& $27.3^{+1.8}$  & $0.4^{+0.3}$ & $3.8^{+1.7}$  \\
	FR*  & 	 $23.4^{+1.1}$	 &  $41.4^{+3.4}$	 & $\textbf{4.8}^{+4.7}$ & $11.7^{+5.5}$    \\
	DR*  &  $\textbf{27.4}^{+1.7}$  & $\textbf{45.3}^{+3.6}$ &  $4.5^{+4.0}$ & $\textbf{12.9}^{+5.3}$\\
	\bottomrule
    \end{tabular}
    \label{visdrone}
\end{minipage}
\begin{minipage}[]{0.48\textwidth}
    \centering
    \caption{Results on DOTA-v2.0. The {\tt train}, {\tt val} sets are used for training and validation. FR, DR denote Faster R-CNN, DetectoRS, * means \textit{with RFLA}.}
    \begin{tabular}{l|cc|cc}  
	\toprule
	Method  & AP & $\rm{AP_{0.5}}$ &  $\rm{AP_{vt}}$ & $\rm{AP_{t}}$  \\
	\midrule
	FCOS     &  31.8	& 55.4	& 0.3 & 4.0  \\
	FR    &  	 35.6	& 59.5 & 0.0 & 7.1    \\
	DR    &  40.8  & 62.6  & 0.0  & 7.0 \\
	\midrule
	FCOS*    & 	 	$32.1^{+0.3}$	& $55.6^{+0.2}$	 & $0.7^{+0.4}$ &  $6.8^{+2.8}$   \\
	FR*   & 	$36.3^{+0.7}$  & $61.5^{+2.0}$ & $1.9^{+1.9}$ & $\textbf{11.7}^{+4.6}$    \\
	DR*  &  $\textbf{41.3}^{+0.5}$  &  $\textbf{64.2}^{+1.6}$  & $\textbf{2.1}^{+2.1}$  & $10.8^{+3.8}$\\
	\bottomrule
    \end{tabular}
    \label{dotav2}
\end{minipage}
\end{table}

\subsection{Analysis}
\label{sec.analysis}

We conduct a group of analysis experiments to delve into different prior designs and assigners for tiny objects. In the first step, we respectively deploy the way of prior tiling in Faster R-CNN ~\cite{Faster-R-CNN_2015_NIPS}, FCOS~\cite{FCOS_2019_ICCV} and RFLA. In the second step, we randomly generate different \textit{gt} in different locations of the image and simulate the process of label assignment for statistics. Concretely, the \gt \ scales are randomly picked from 0 to 64. After that, we divide the scale range into 16 intervals, as shown in Fig.~\ref{fig:imbalance}, and calculate the average number of positive samples assigned to each \gt \ in different scale ranges. Observations in Fig.~\ref{fig:imbalance} indicate severe scale-sample imbalance problems for existing detectors. For anchor-based detectors, objects in the tiny scale and the interval between box scales become outliers. Anchor-free detectors somewhat alleviate this problem. However, tiny objects are still outliers since tiny object covers an extremely limited region. The number of prior points inside \gt \ is much smaller than that of large objects. The scale-sample imbalance problem will mislead the network towards unbalanced optimization, where less focus is imposed on outlier samples. In contrast, the number of positive samples assigned to \gt \ in different scale ranges is greatly reconciled with RFLA, achieving a balanced optimization for tiny objects.

\subsection{Experiment on more datasets}
\begin{figure*}[h]
    \renewcommand{\arraystretch}{0.8} 
    \renewcommand{\tabcolsep}{0.45mm}
    \newcommand{\rowwidth}{2.7cm}
    \centering
    \begin{tabular}{C{\rowwidth}C{\rowwidth}C{\rowwidth}C{\rowwidth}}
         \includegraphics[width=1.0\linewidth]{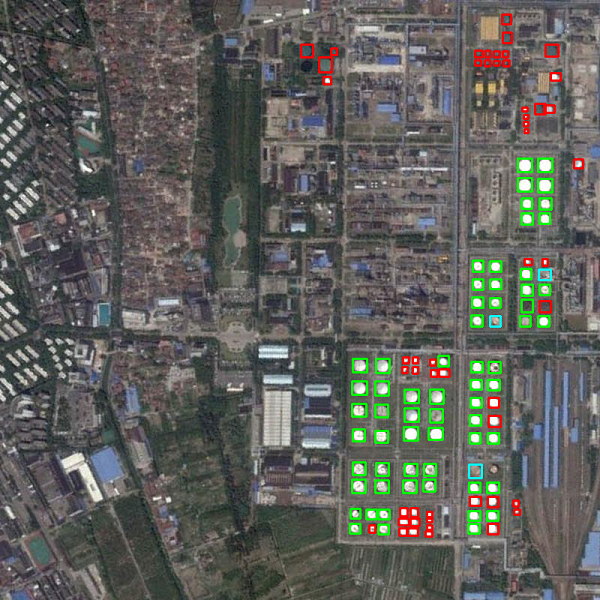} &
         \includegraphics[width=1.0\linewidth]{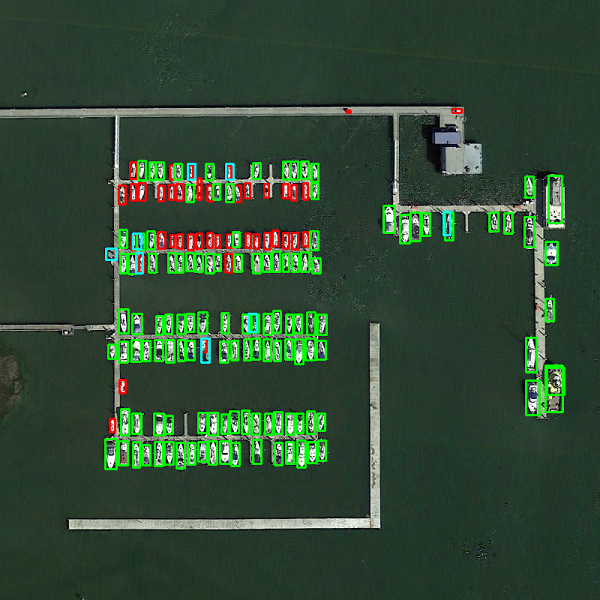} &
         \includegraphics[width=1.0\linewidth]{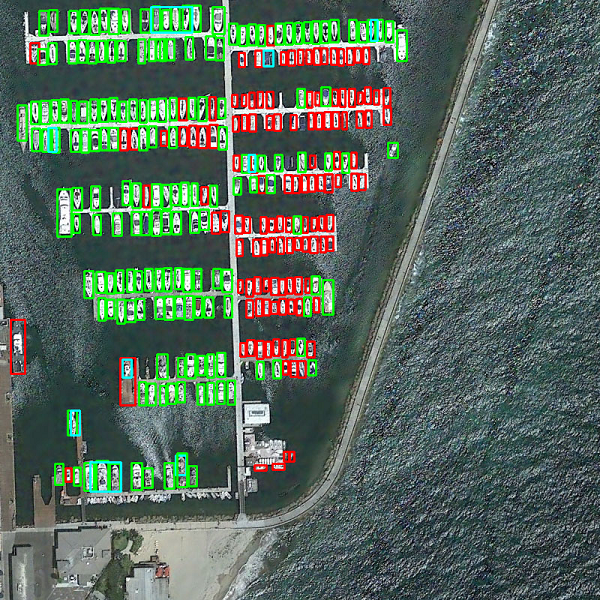} &
         \includegraphics[width=1.0\linewidth]{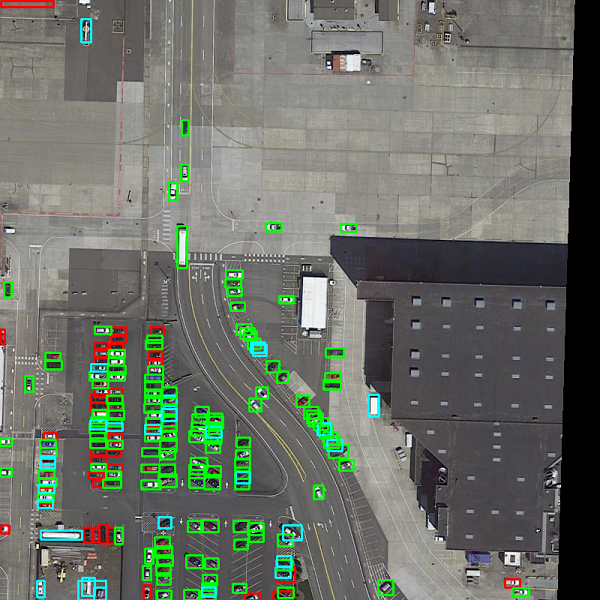} \\
         \includegraphics[width=1.0\linewidth]{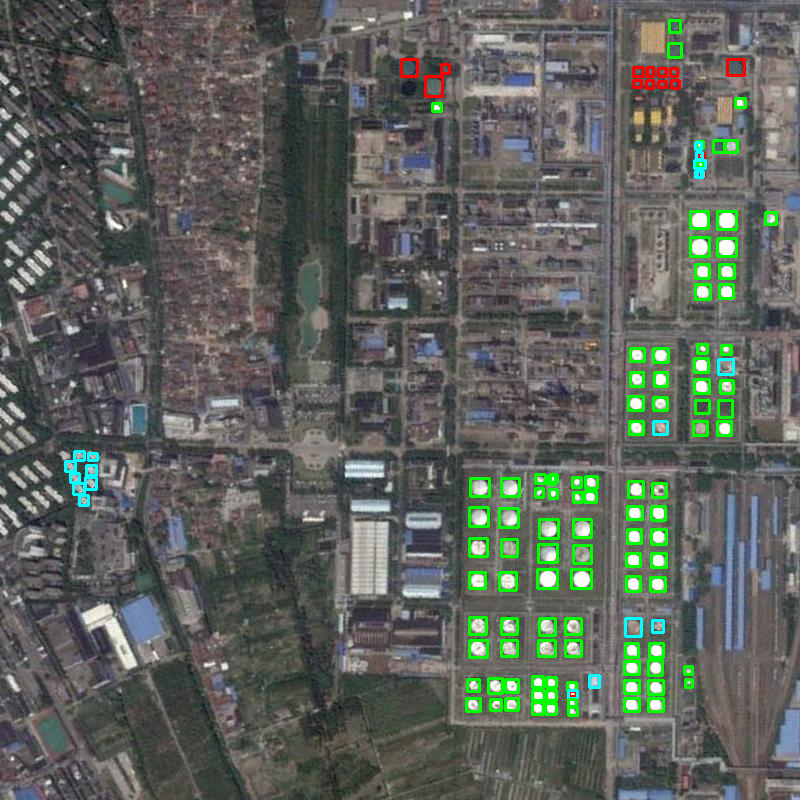} &
         \includegraphics[width=1.0\linewidth]{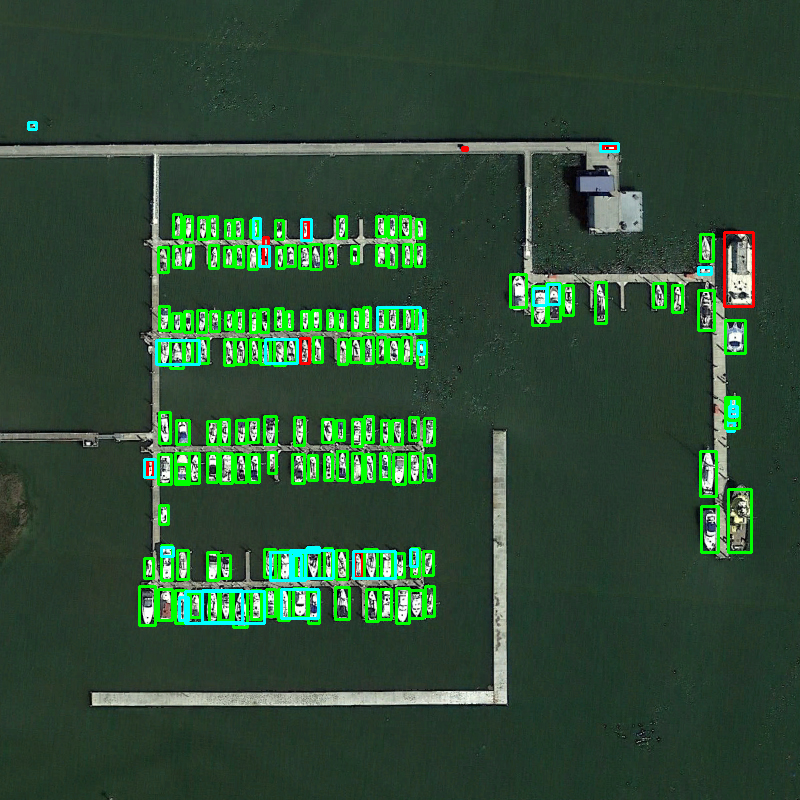} &
         \includegraphics[width=1.0\linewidth]{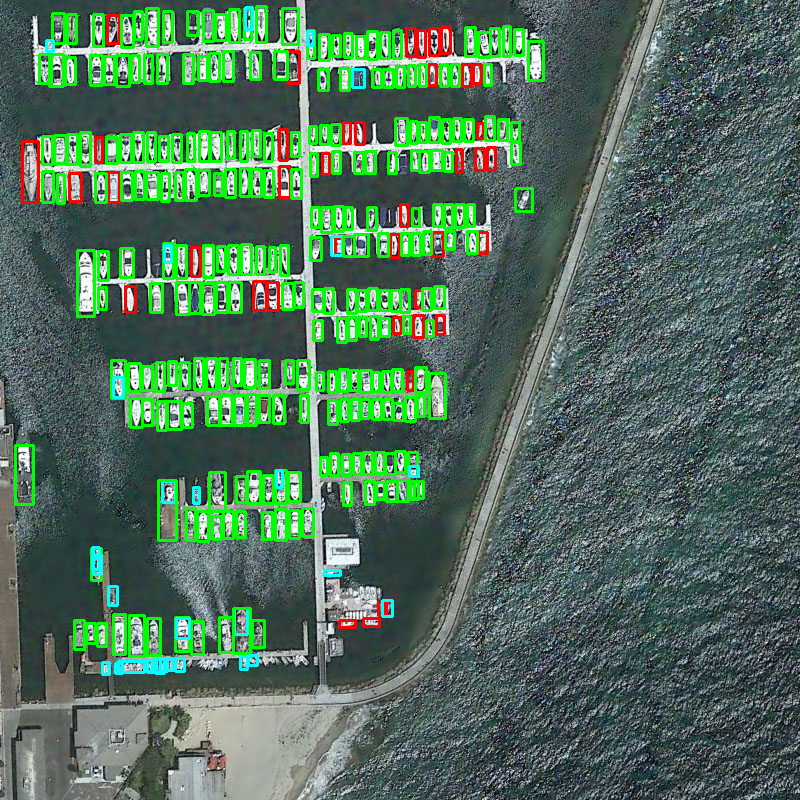} &
         \includegraphics[width=1.0\linewidth]{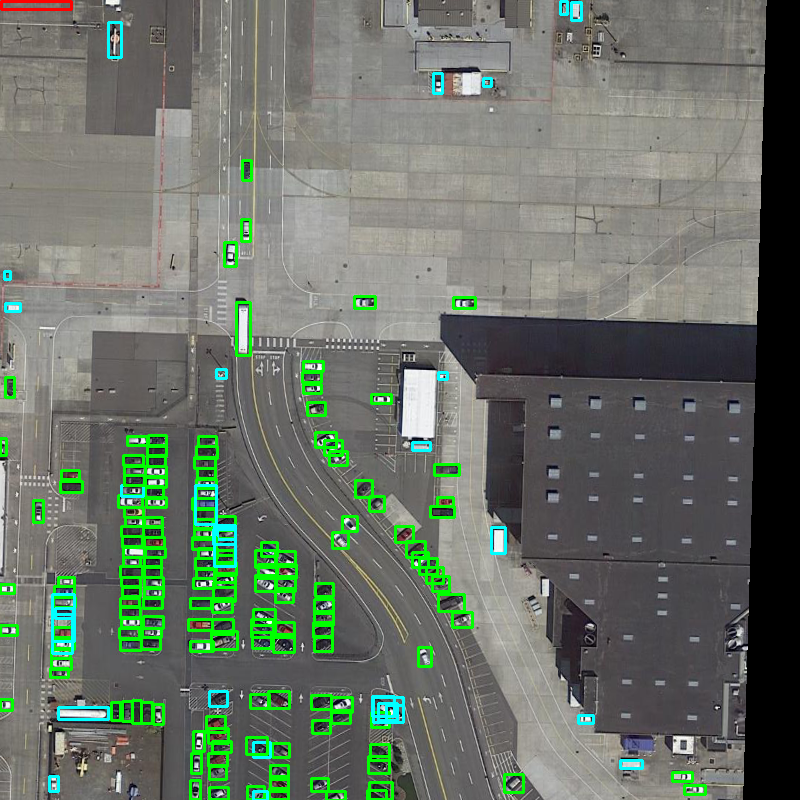}
    \end{tabular}
    \caption{Visualization results on AI-TOD. The first row is the result of Faster R-CNN and the second row is the result of Faster R-CNN w/ RFLA.}
    \label{fig:vis}
\end{figure*}

We conduct experiments on another TOD dataset TinyPerson~\cite{TinyPerson_2020_WACV}. The dataset setting and evaluation all follow TinyPerson benchmark~\cite{TinyPerson_2020_WACV}. The results are in Tab.~\ref{table:tinyperson}, 3.1 and 1.4 $\rm{AP_{50}^{tiny}}$ improvement can be obtained when applying RFLA into FCOS and Faster R-CNN. We also tested the RFLA on AI-TOD-v2~\cite{aitodv2_2022_isprs}, whose performance is listed in \href{https://github.com/Chasel-Tsui/mmdet-rfla}{GitHub}. 
We further verify the effectiveness of RFLA on datasets which simultaneously hold a large scale variance and contain many tiny objects (\textit{i.e.}~VisDrone2019, DOTA-v2.0). The results are in Tab.~\ref{visdrone} and Tab.~\ref{dotav2}. The improvement of  $\rm{AP_{vt}}$ and $\rm{AP_{t}}$ is quite obvious for both datasets. The consistent improvement on various datasets indicates RFLA's generality. Finally, visualization results on the AI-TOD dataset are shown in Fig.~\ref{fig:vis}. When applying RFLA into Faster R-CNN, FN predictions can be greatly eliminated.

\section{Conclusion}

In this paper, we point out that box and point prior do not work well for TOD, leading to scale-sample imbalance problems when assigning labels. To this end, we introduce a new Gaussian receptive field prior. Then, we further design a new Receptive Field Distance (RFD), which measures the similarity between ERF and \gt~to conquer the shortages of IoU and centerness on TOD. 
The RFD works with the HLA strategy, obtaining balanced learning for tiny objects. Experiments on four datasets show the superiority and robustness of the RFLA.

\section*{Acknowledgement}

This work was partly supported by the Fundamental Research Funds for the Central Universities under Grant 2042022kf1010, and the National Natural Science Foundation of China under Grant 61771351 and 61871297. The numerical calculations were conducted on the supercomputing system in the Supercomputing Center, Wuhan University.

%
%
\bibliographystyle{splncs04}
\bibliography{chang, Jinwang}
\end{document}